\documentclass{article}
\usepackage{ijcai20}

\usepackage{times}
\usepackage{soul}
\usepackage{url}
\usepackage{color}
\usepackage[hidelinks]{hyperref}
\usepackage[utf8]{inputenc}
\usepackage[small]{caption}
\usepackage{graphicx}
\usepackage{amsmath}
\usepackage{amsthm}
\usepackage{booktabs}
\usepackage{algorithm}
\usepackage{algorithmic}
\usepackage{multirow}
\usepackage{setspace}

\title{Robust Person Re-Identification through Contextual Mutual Boosting}
\author{
Zhikang Wang$^{1,2}$\footnote{Contact Author}\and
Lihuo He$^1$\and
Xinbo Gao$^1$\And
Jane Shen$^2$\\
\affiliations
$^1$Xidian University\\
$^2$Pensees Singapore\\
\emails
\{zkwang00, lihuo.he\}@gmail.com,
xbgao@mail.xidian.edu.cn,
jane.shen@pensees.ai
}

\begin{document}
	
	\maketitle
	\begin{abstract}
		Person Re-Identification (Re-ID) has witnessed great advance, driven by the development of deep learning. However, modern person Re-ID is still challenged by background clutter, occlusion and large posture variation which are common in practice. Previous methods tackle these challenges by localizing pedestrians through external cues (e.g., pose estimation, human parsing) or attention mechanism, suffering from high computation cost and increased model complexity. In this paper, we propose the Contextual Mutual Boosting Network (CMBN). It localizes pedestrians and recalibrates features by effectively exploiting contextual information and statistical inference. Firstly, we construct two branches with a shared convolutional frontend to learn the foreground and background features respectively. By enabling interaction between these two branches, they boost the accuracy of the spatial localization mutually. Secondly, starting from a statistical perspective, we propose the Mask Generator that exploits the activation distribution of the transformation matrix for generating the static channel mask to the representations. The mask recalibrates the features to amplify the valuable characteristics and diminish the noise. Finally, we propose the Contextual-Detachment Strategy to optimize the two branches jointly and independently, which further enhances the localization precision. Experiments on the benchmarks demonstrate the superiority of the architecture compared the state-of-the-art.
	\end{abstract}

	\begin{figure*}[htb]
		\centerline{\includegraphics[width=15cm, height=6cm]{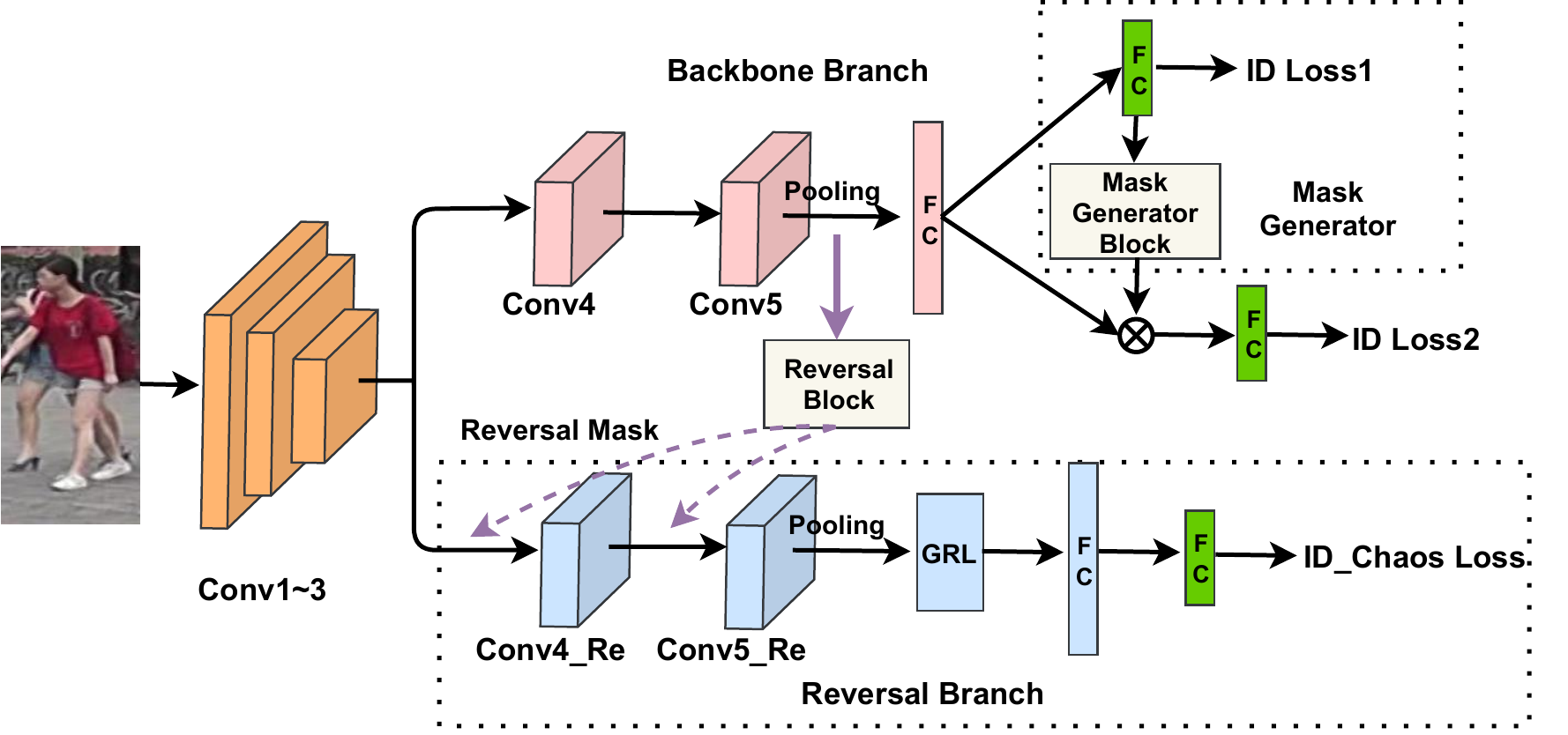}}
		\caption{The overall architecture of our proposed Contextual Mutual Boosting Network (CMBN). There are four components, i.e., the shared convolutional frontend, the Backbone Branch, the Reversal Branch, and the Mask Generator Model. GRL and FC refer to the Gradient Reversal Layer and Fully-Connected Layer, respectively. During training, three ID losses will jointly optimize the whole network. In the inference phase, to be computationally efficient, the Reversal Branch and the Mask Generator will be discarded. The features multiplied by the generated mask form the final representation.}
	\end{figure*}
	
	\section{Introduction}
	Person Re-Identification (Re-ID) aims at retrieving images of the same person as a given query image from a   `gallery' set of images. Numerous methods \cite{qian2018pose,si2018dual,sun2018beyond,su2017pose,song2018mask} have been proposed to address the problem due to the significant impact of Re-ID in real-life application.
	However, existing methods typically suffer from performance degradation due to challenging factors like occlusion, background clutter, and posture changes. Therefore, there is consequently an inevitable need for training a robust model that can localize and recognize the pedestrians accurately.

	Localizing pedestrians in an image is an effective way to alleviate negative effects from the above distracting factors. Motivated by the idea, numerous methods \cite{zheng2019re,song2018mask,zhao2017spindle,kalayeh2018human,qi2018maskreid,liu2018pose} have been developed to generate discriminative and robust person representations.
	Among them, some methods \cite{liu2018pose,cho2016improving} perform pose estimation at first for inferring the location of the pedestrians indirectly. Some other methods \cite{kalayeh2018human,song2018mask}  adopt auxiliary algorithms to segment pedestrians from the images directly. Differently, the attention mechanism, which is capable of localizing the informative parts of an input image, is also adopted for coarsely localizing the pedestrians and filtering distracting factors  \cite{li2018harmonious}. However, all these methods suffer from a  common drawback, i.e., the high computation cost for both the training and inference phases. Besides, the performance of those methods that require external clues (e.g., pose, segmentation) relies heavily on the accuracy of those external clues, hurting the robustness of the overall model.

    Recalibrating the features is also effective in improving the discrimination of the final representations for recognition. As each feature channel is considered as a specific feature detector \cite{zeiler2014visualizing}, recalibrating the attention weights on the channel dimension can dilute the noise and amplify the valuable characteristics. Quite a few methods \cite{li2018harmonious,cbam,hu2018squeeze} focus on this line. However, they all follow the `squeeze and excitation' mode \cite{hu2018squeeze}, that first produces the channel descriptor through a squeeze operation and then produces a collection of channel modulation weights from the embeddings through an excitation operation. The limitation is that only stacking many such blocks can bring significant improvements to the discrimination, incurring high computation overhead.

	In this paper, we propose the Contextual Mutual Boosting Network (CMBN) to generate more discriminative features for pedestrians by spatial localization and channel recalibration. Specifically, we construct the Backbone Branch (BB) and the Reversal Branch (RB) to learn the ID-discriminative foreground and ID-irrelevant background features of images, respectively. By establishing the interactions, the two branches boost the corresponding spatial localization mutually during the training phase. Secondly, we propose the Mask Generator (MG) to recalibrate the extracted features. It aims to generate a universal and static channel mask, which amplifies the valuable characteristics and diminishes the noise, by exploiting the activation distribution of the transformation matrix. 
	Thirdly, we propose the Contextual-Detachment Strategy (CDS), which further improves the spatial localization precision by thining the optimization process of the two branches.
	Both of the RB and MG are auxiliary components, meaning that they will be discarded after training and only the generated channel mask is needed for multiplication. Therefore, our methods can generate discriminative representations with little additional computation overhead.

	In summary, the contributions of this paper are as follow:
	\begin{itemize}
		\item We propose the Contextual Mutual Boosting Network (CMBN), which localizes pedestrians and recalibrates features by exploiting the contextual information and statistical inference, respectively.
		\item We propose a novel Reversal Branch, which aims at extracting the background information of images to boost the pedestrians' spatial localization indirectly.
		\item We propose a novel Mask Generator to learn a universal and static channel mask that will recalibrate the features on the channel dimension to amplify the valuable characteristics and diminish the noise.
		\item We propose a Contextual-Detachment Strategy to optimize the two branches with contrary functions. It helps the Backbone Branch with better foreground sensitivity and more accurate spatial localization.
	\end{itemize}

    To demonstrate the effectiveness of our approach, we conduct experiments on three challenging Re-ID datasets. Our method achieves competitive performance under multiple evaluation metrics.

	\section{Related Work}
	\subsubsection{Deep Learning Based Person Re-ID}
	The deep learning has witnessed great success in person Re-ID. Many works emerge after deep learning surpassing human-level performance in image classification tasks like ImageNet \cite{imagenet_cvpr09}. Meanwhile, many large scale person Re-ID datasets that enable training of deep networks become available, for instance Market1501 \cite{zheng2015scalable}, CUHK03 \cite{li2014deepreid} and DukeMTMC-ReID\cite{dukemtmc}, which accelerate the development in this area. In \cite{zheng2016person}, an identification network is proposed for tackling person Re-ID. Afterwards, some domain-specific designs emerged, like PCB\cite{sun2018beyond}, MGN\cite{mgn}. 
	
	\subsubsection{Human Body Information}
	In this work, we attempt to generate representations robust to occlusion, posture variations, and background clutter. Similar approaches make use of body key points, parts region, human parsing or segmentation. In \cite{zhao2017spindle}, the authors apply a coarse-to-fine region feature extraction and merging process to generate robust and discriminative person representations. In \cite{miao2019pose}, a similar divide-then-merge method are exploited. The authors use key points to generate attention and guide the network to focus on areas that can be seen so that occluded parts are not influencing the representation. 
	In \cite{qi2018maskreid}, the authors propose to use human body segmentation mask to guide the network to learn robust features.  Moreover, a fine-grained human body segmentation is proposed in \cite{hsp} to generate features for each semantic part. 
	
	\subsubsection{Attentions}
	Those aforementioned methods utilize auxiliary modules with fixed parameters. Such configurations are sub-optimal because of mismatch between the domains of the main model and the auxiliary model. Hence, end-to-end-trainable modules are in more favor (than those require multiple training stages), such as attention modules. Here, we elaborate on two kinds of attention mechanisms: spatial attention and channel attention. 
	
	Spatial attention is useful for allocating available resources towards the informative parts of images. For example, \cite{sa} introduced spatial self attention on the feature maps before the Global Average Pooling. \cite{xu2018attention} aggregates the pose estimation with the spatial attention together for better utilization of the pose information.
	Channel attention is mainly for features recalibration. In \cite{hu2018squeeze}, the `squeeze-and-excitation' block is proposed. It firstly aggregates the feature maps across the spatial dimensions and then passes them through a squeeze operation for producing the channel descriptor. The squeeze features are embeddings of the global distribution on the channel dimension. Then, an excitation operation, which takes the embeddings as input and produces a collection of channel modulation weights, is followed. 
	
	There are also works that make effort in utilizing both kinds of attention mechanisms. For instance: \cite{cbam}, both spatial and channel attention are introduced to enhance the performance of ResNet \cite{resnet} architectures. \cite{mhoa} generalizes attention masks from spatial attention and channel attention to attention maps with high-order statistics information. \cite{li2018harmonious} propose the harmonious attention module that simultaneously learns `hard' region level and `soft' pixel-level attentive features for multi-granular feature representation.

	\section{The Proposed Method}
	In this section, we first introduce the architecture of our proposed Contextual Mutual Boosting Network (CMBN) and then elaborate on each of its components. As illustrated in Figure 1, apart from the shared convolutional frontend, there are two branches with different learning targets.  
	The Backbone Branch (BB) (Sec.3.1) aims to learn the ID-discriminative foreground features of pedestrians for identification and the Reversal Branch (RB) (Sec.3.2) extracts the ID-irrelevant background features for boosting the BB's pedestrian spatial localization indirectly. Meanwhile, at the end of the Backbone Branch, we propose the Mask Generator (MG) (Sec.3.3) to generate the static and universal channel mask for recalibrating the final representations.
	During the inference stage, both the RB and MG will be discarded. We regard the BB's features multiplied by the channel mask as the final representation. 
	
	%For the upper branch, it aims at learning the discriminative features of the corresponding images. As for the other one, it learns the comtaminated factors like occlusion and background clutter, which spatial localization of pedestrians indirectly. 
	
	\subsection{The Backbone Network}
	In this paper, we adopt the ResNet50 \cite{resnet}, which is well-known to extract the spatial features, as the backbone network. 
	It consists of 1 convolutional block and 4 residual convolutional blocks named $Conv1\sim5$ respectively. 
	Here, we make three modifications on the original network: (1) we replace the Global Average Pooling (GAP) by the Global Maximum Pooling (2) we discard the final category classification layer, which is for classification on the ImageNet dataset \cite{imagenet_cvpr09}, and add another two fully connected layers for producing the identity prediction of the input images (3) we remove the last spatial down-sampling to increase the granularity of features. The combination of the backbone network and FC layers is the IDE baseline \cite{zheng2016person}.
	%The Backbone network is the IDE baseline \cite{zheng2016person}.

	\begin{figure}[htb]
		\centerline{\includegraphics[width=8cm, height=2.5cm]{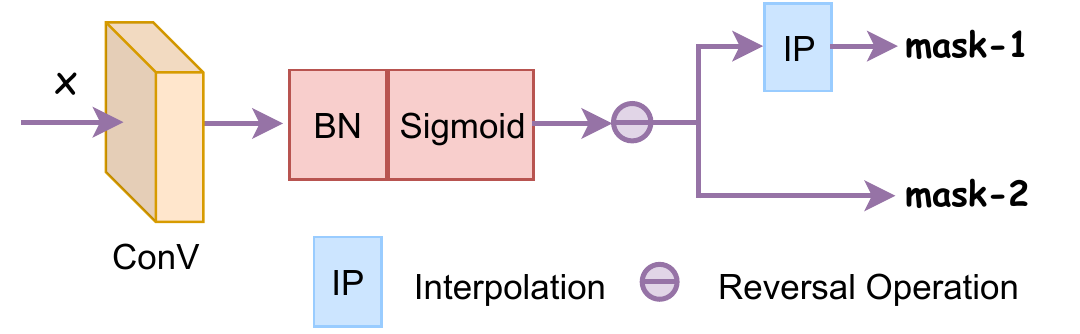}}
		\caption{Illustration of the Reversal Block.}
	\end{figure}

	\subsection{Reversal Branch}
	We aim to achieve the spatial localization function by the mutual boosting of the multiple contextual information.
	To achieve the goal, we propose the Reversal Branch (RB), which is the combination of the Convolutional Blocks and the Gradient Reversal Layer (GRL) \cite{ganin2014unsupervised}, for learning the ID-irrelevant background features.
	It is known that GRL successfully combines domain adaptation and deep feature learning within one training process (deep domain adaptation). 
	By maximizing the loss of the domain classifier, the network can not identify the domain categories correctly and thus generates the domain-invariant features. Following the idea, we introduce the GRL for enforcing the Reversal Branch to extract the ID-irrelevant background features, which are not optimal for identification, by maximizing the ID classification loss. %We name it as the ID\_Chaos Loss ($L_{CS}$) and the function can be formulated as:
	More formally, the function can be formulated as:
	\begin{equation}
	E({\theta _{id}},f) =  - \lambda \sum\limits_{i = 1...N} {{L_y}({G_{id}}(f,{\theta _{id}}),y)},
	\end{equation}
	where $L_{y}$ is the Cross Entropy Loss, $f$ is the feature representation of the images, $y$ is the corresponding label, $G_{id}$ is the ID classifier, and $\theta _{id}$ is classifier parameters.

	As shown in Figure 1, features extracted from the shared convolutional frontend $Conv1\sim3 $, will be sent into the BB and RB simultaneously. For the BB, features will go through the $Conv4\sim5$. The obtained feature maps have two functions: (1) training the network's pedestrian identification ability by minimizing the ID Loss (2) generating the reversal mask, whose spatial area roughly covers the ID-irrelevant background regions, for constraining the RB. 
	To achieve the second function, we insert the Reversal Block, which is shown in Figure 2, between the BB and the RB. For the Reversal Block, it first adopts a learnable convolutional layer with $1 \times 1$ kernel size for aggregating features along the channel dimension. A Batch Normalization (BN) is followed. Next, we apply the non-linear function $Sigmoid$ for normalizing the value between 0 and 1. 
	Then, we conduct the Reversal Operation by subtracting the value-limited output from 1 for getting the reversal mask. The function can be formulated as:
	\begin{equation}
	h=1-Sigmoid(Conv(fm)),
	\end{equation}
	where $fm$ represents the feature maps from the BB, $Conv$ is the convolutional layer, $h$ is the generated mask.
After getting the mask, an interpolation with a scale factor as 2 is adopted for enlarging the spatial size. The enlarged and original maps will be multiplied on the inputs of the $Conv4\_{Re}$ and $Conv5\_{Re}$ respectively for enforcing the spatial constrains. After going through all the $Conv\_{Re}$ blocks, the GRL with hyperparameter $\lambda$ as 0.2 is inserted. Then, two FC layers for identity prediction is subsequent. During the forward propagation, the BB helps to shape the background spatial region of RB through the reversal masks. During the backpropagation, in verse, the RB helps to shape the foreground spatial region of BB by propagating the gradients through the Reversal Block. In this way, we achieve a mutual boosting scheme.
The ID Loss will sustain a high value during the training process since the extracted features, which focus on the background information, lack the ID-relevant characteristics.

	\subsection{Mask Generator}
	Recalibration plays an important role in feature refinement. Previous works, including image classification \cite{cbam,hu2018squeeze}, person Re-ID \cite{li2018harmonious,mhoa} have widely adopted the attention mechanism for features recalibration. As discussed previously, many stacked SEblocks are required for noticeable improvement of the feature representation.
    To this end, we proposed a novel Mask Generator (MG), which aims to learn a static and universal mask for all the representations directly.

	As we know, person Re-ID is a task that combines classification and retrieval together, meaning that no overlapping identities between training and testing set. We optimize the network by an additional FC layer, parametrized by a matrix $W$, for ID classification. In the training phase, the FC layer fits the data well and achieves 100\% accuracy quickly. From the perspective of statistical inference, the activation distribution of the transformation matrix reflects the value of each channel indirectly.
	Our MG dedicates to exploring the statistical information for feature recalibration.

	As shown in Figure 1, we achieve the function by adding a FC layer and a mask generator block. 
	Specifically, we first adopt an additional FC layer, parameterized by a weight matrix, $W^{ID}$, for ID classification. The weight matrix and former network will be optimized by an ID Loss1. Then, we input the matrix $W^{ID}$ into the Mask Generator Block for generating the final mask. The block is constructed by another FC layer, parametrized by the weight matrix, $W^G$, and a $Sigmoid$ function.
	The matrix $W^G$ will generate a weight for each channel after inferring all the training identities' feature activation distribution and $Sigmoid$ is for normalizing the weight to 0 and 1.
	The function can be formulated as:
	\begin{equation}
	Mask = Sigmoid({W^G}({W^{ID}})).
	\end{equation}
	Since the generated mask will be multiplied on the representations, the MG will be optimized by the ID Loss2. The generated mask is not dynamically varying with the input, hence at the end of the training, there is no need to reserve the MG, we can only preserve the channel mask for recalibration through multiplication.

	\begin{algorithm}[htb] 
		\caption{Contextual-Detachment Strategy.} 
		\label{alg:Framwork} 
		\begin{algorithmic}[1] %这个1 表示每一行都显示数字
			\renewcommand{\algorithmicrequire}{\textbf{Input:}}
			\renewcommand{\algorithmicensure}{\textbf{Output:}}
			\REQUIRE ~~\\ %算法的输入参数：Input
			A mini-batch sample images of arbitrary identities;
			\ENSURE ~~\\ %算法的输出：Output
			CMBN;
			\STATE \textbf{while} not converge \textbf{do}; 
			\STATE  Optimizing the whole network with the two ID Losses and one ID\_Chaos Loss: \\
			\centerline{$L=L_{ID1} + L_{ID2} + \beta L_{CS}$}
			\STATE  Fine-tuning the shared convolutional frontend and Backbone Branch with the two ID losses: \\
			\centerline{$L=L_{ID1} + L_{ID2}$}
			\STATE \textbf{end while};
		\end{algorithmic}
	\end{algorithm}

	\subsection{Contextual-Detachment Strategy}
	As illustrated in Figure 1, two ID Losses ($L_{ID1}$ and $L_{ID2}$) and one ID\_Chaos Loss ($L_{CS}$) are used for optimizing the whole network. These three losses are all Cross Entropy Losses for classification. Note that the first two ID losses are applied to learn the foreground features, while the third one enables the Reversal Branch to focus on the ID-irrelevant background information.
	
	\begin{figure}[htb]
		\centerline{\includegraphics[width=7cm, height=2cm]{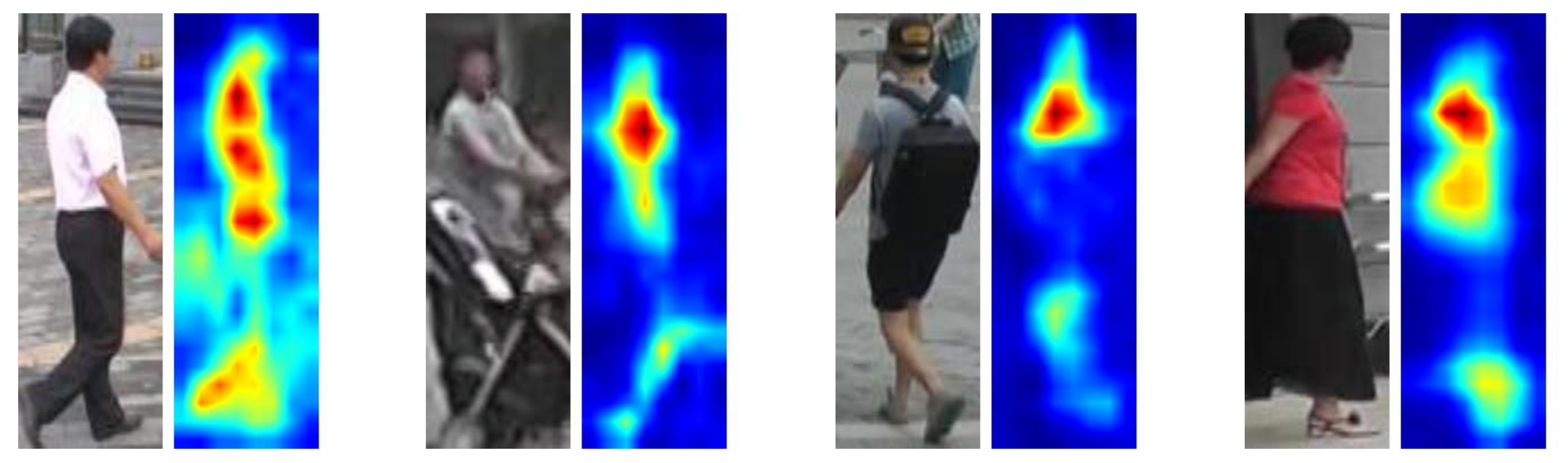}}
		\caption{Visualization of the Backbone Branch's activation maps.}
		\label{fig3}
	\end{figure}
	
	A straight forward idea is to jointly train the network with the three losses. To balance the conflicts of the two branches, a hyperparameter $\beta$ is multiplied on the $L_{CS}$. Here, we set $\beta$ as 0.4.
	In Figure \ref{fig3}, we visualize the activation maps of the Backbone Branch's outputs, which is optimized through the training strategy. We find that the background and occlusion pollute the feature maps significantly and some valuable characteristics are not salient enough. We argue that the problems are caused by the disharmonious cooperation between the two branches.
	Therefore, we propose the Contextual-Detachment Strategy (CDS), which is shown in Alg.\ref{alg:Framwork}, to optimize the network.
	Specifically, we first jointly update the two branches with the three losses, which allows the two branches to focus on their corresponding contextual information and generate coarse activation maps. Then, we adopt only the two ID Losses to fine-tune the Backbone Branch. Through the training strategy, the BB can generate better-refined foreground features. In reverse, better foreground spatial localization boosts the background characteristics learning through the Reversal Block. The alternately optimization process is conducted end-to-end.

	\begin{figure*}[htb]
		\centerline{\includegraphics[width=17cm, height=4.5cm]{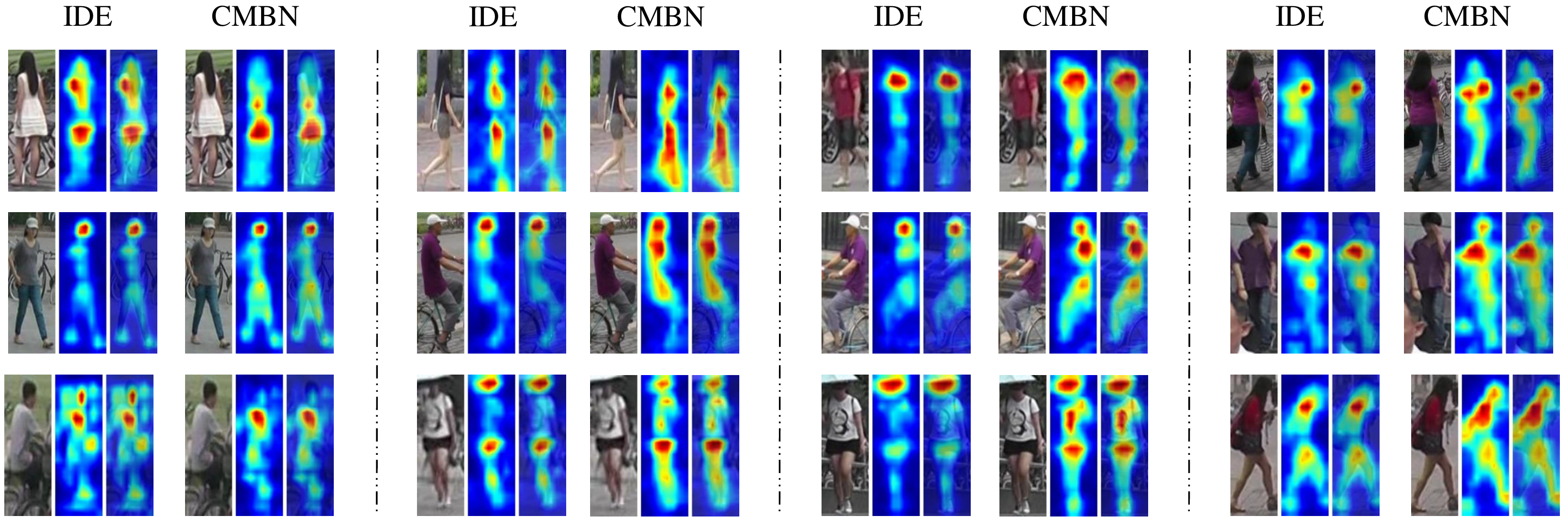}}
		\caption{Visualization of the activation maps of the IDE and CMBN respectively. For better visual effects, we also map the activation maps on the original images.}
	\end{figure*}

	\begin{table}[t]
		\centering
		\caption{Results over Market-1501, Single Query (SQ)}\smallskip
		\setlength{\tabcolsep}{2mm}{
			\begin{tabular}{cc|cc}
				\hline
				\multicolumn{4}{l}{Market-1501} \\
				\hline
				\multicolumn{2}{l|}{Methods} & R=1 & mAP \\
				\hline
				\hline
				\multicolumn{2}{l|}{BoW+kissme\cite{zheng2015scalable}} & 44.4 & 20.8 \\
				\multicolumn{2}{l|}{LOMO+XQDA\cite{liao2015person}}& 43.8 & 22.2   \\
				\hline
				
				\multicolumn{2}{l|}{MGCAM\cite{song2018mask}}& 83.79 & 74.33   \\
				\multicolumn{2}{l|}{SPReID\cite{kalayeh2018human}}& \textbf{93.68} & \textbf{83.36}   \\

				\hline
				\multicolumn{2}{l|}{PDC\cite{su2017pose}}& 84.14 & 63.41 \\
				\multicolumn{2}{l|}{AACN\cite{xu2018attention}}& 85.90 & 66.87 \\
				\multicolumn{2}{l|}{PN-GAN\cite{qian2018pose}}& 89.43 & 72.58 \\

				\multicolumn{2}{l|}{PGFA\cite{miao2019pose}}& 91.2 & 76.8 \\
				
				\hline
				\multicolumn{2}{l|}{HA-CNN\cite{li2018harmonious}} & 91.2& 75.7 \\
				\multicolumn{2}{l|}{DuATM\cite{si2018dual}} & 91.4& 76.6 \\
				\multicolumn{2}{l|}{CASN(IDE)\cite{zheng2019re}}  & 92.0 & 78.0 \\
				
				\hline
				\hline
				\multicolumn{2}{l|}{CMBN} & 92.8 &  79.8 \\
				\hline
		\end{tabular}}
	\end{table}
	
	\begin{table}[t]
		\centering
		\caption{Results over DukeMTMC-ReID, SQ}\smallskip
		\setlength{\tabcolsep}{2mm}{
			\begin{tabular}{cc|cc}
				\hline
				\multicolumn{4}{l}{DukeMTMC-ReID} \\
				\hline
				\multicolumn{2}{l|}{Methods} & R=1 & mAP \\
				\hline
				\hline
				\multicolumn{2}{l|}{BoW+kissme\cite{zheng2015scalable}} & 25.1 & 12.2 \\
				\multicolumn{2}{l|}{LOMO+XQDA\cite{liao2015person}}& 30.8 & 17.0   \\
				
				\hline
				\multicolumn{2}{l|}{SPReID\cite{kalayeh2018human}}& \textbf{85.95} & \textbf{73.34} \\
				\hline
				
				\multicolumn{2}{l|}{PN-GAN\cite{qian2018pose}}& 73.58 & 53.20 \\
				\multicolumn{2}{l|}{AACN\cite{xu2018attention}}& 76.84 & 59.25 \\
				\multicolumn{2}{l|}{PGFA\cite{miao2019pose}}& 82.6 & 65.5 \\
				
				\hline
				\multicolumn{2}{l|}{HA-CNN\cite{li2018harmonious}} & 80.5& 63.8 \\
				\multicolumn{2}{l|}{DuATM\cite{si2018dual}} & 81.8& 64.6 \\
				\multicolumn{2}{l|}{CASN(IDE)\cite{zheng2019re}}  &84.5 & 67.0 \\
				
				%\multicolumn{2}{l|}{SVDNet\cite{sun2017svdnet}} & 82.3 & 62.1 \\
				%\multicolumn{2}{l|}{MLFN} & 90.0 & 74.3 \\
				%\multicolumn{2}{l|}{KPM} & 90.1 & 75.3 \\
				%\multicolumn{2}{l|}{DNN-CRF} &  93.5  & 81.6 \\
				%\multicolumn{2}{l|}{PABR} &  91.7  & 79.6 \\
				%\multicolumn{2}{l|}{PCB+RPP\cite{sun2018beyond}} &  93.8 & 81.6 \\
				
				\hline
				\hline
				\multicolumn{2}{l|}{CMBN} & 84.8 &  69.5 \\
				\hline
		\end{tabular}}
	\end{table}
	
	\begin{table}[t]
		\centering
		\caption{Results comparisons over CUHK03-NP, SQ}\smallskip
		\setlength{\tabcolsep}{0.5mm}{
			\begin{tabular}{cc|cc|cc}
				\hline
				\multicolumn{5}{l}{CUHK03-NP} \\
				\hline
				\multirow{2}{*}{Methods} &  & \multicolumn{2}{l|}{Detected} &  \multicolumn{2}{l}{Labeled}\\
				~ & ~ & R-1 & mAP & R-1  & mAP \\
				\hline
				\hline
				\multicolumn{2}{l|}{BoW+XQDA\cite{zheng2015scalable}} & 6.4 & 6.4 & 7.9 & 7.3 \\
				\multicolumn{2}{l|}{LOMO+XQDA\cite{liao2015person}}& 12.8 & 11.5 & 14.8 & 13.6 \\
				\hline
				
				\multicolumn{2}{l|}{MGCAM\cite{song2018mask}}& 46.71 & 46.87 & 50.14 & 50.21   \\
				
				\hline
				\multicolumn{2}{l|}{HA-CNN\cite{li2018harmonious}} & 41.7& 38.6 & 44.4 & 41.0 \\
				
				\multicolumn{2}{l|}{CASN(IDE)\cite{zheng2019re}}  & 57.4 & 50.7 & 58.9 & 52.2 \\
				\hline
				\hline
				\multicolumn{2}{l|}{CMBN} & \textbf{63.1} & \textbf{58.9}  & \textbf{67.5} & \textbf{63.0} \\
				\hline
		\end{tabular}}	
	\end{table}
	
	\section{Experiments}
	\subsection{Datasets and Evaluation Measures}
	\textbf{Datasets.} In this paper, we adopt the Market-1501 \cite{zheng2015scalable}, CUHK03-NP \cite{li2014deepreid} and DukeMTMC-ReID\cite{dukemtmc} datasets for experiments. Market-1501 collects totally 12,936 training images across 6 cameras without overlapping views. As for the testing data, gallery and query sets consist of 19,732 and 3,368 images respectively with 750 different identities. CUHK03-NP adopts a new split protocol for the CUHK03. Following the new protocol, there are 767/700 identities for training and testing respectively. DukeMTMC-ReID includes 16,522 training images of 702 identities, 2,228 query and 17,661 gallery images of another 702 identities. 
	
	\noindent\textbf{Evaluation Measures.} Cumulative matching characteristics (CMC) cure and mAP are adopted for evaluation. They present the accuracy of the retrieval and the recall rate, respectively.

	\begin{table*}[t]
		\centering
		\caption{Results comparisons over Market-1501 under single query setting}\smallskip
		\setlength{\tabcolsep}{3mm}{
			
			\begin{tabular}{cc|cc|cc|cc|cc}
				\hline
				\multicolumn{2}{c|}{} & \multicolumn{2}{c|}{}&\multicolumn{2}{c|}{}&\multicolumn{4}{c}{CUHK03-NP} \\
				\cline{7-10}
				\multicolumn{2}{c|}{} & \multicolumn{2}{c|}{Market1501}&\multicolumn{2}{c|}{DukeMTMC-ReID}&\multicolumn{2}{c|}{Detected}&\multicolumn{2}{c}{Labeled} \\	
				\hline
				\multicolumn{2}{l|}{Methods} & Rank-1 & mAP & Rank-1 & mAP & Rank-1 & mAP & Rank-1 & mAP\\
				\hline
				\multicolumn{2}{l|}{BaseLine} & 89.5 & 70.7 & 80.2 & 62.3 & 50.8 & 46.0 & 53.1 & 48.1\\
				\multicolumn{2}{l|}{BaseLine+RB} & 91.4 & 77.8 & 84.1 & 67.5 & 59.4 & 55.1 & 60.4 & 56.9\\
				\multicolumn{2}{l|}{BaseLine+MG} & 91.7 & 78.9 & 84.8 & 69.0 & 60.9 & 57.4 & 64.8 & 61.7\\
				\multicolumn{2}{l|}{BaseLine+RB+MG} & 91.9 & 78.8 & 84.3 & 68.4 & 60.1 & 56.2 & 63.3 & 60.2\\
				\multicolumn{2}{l|}{BaseLine+RB+MG+CDS} & \textbf{92.8} & \textbf{79.8} & \textbf{84.8} & \textbf{69.5} & \textbf{63.1} & \textbf{58.9} & \textbf{67.5} & \textbf{63.0}\\
				\hline
		\end{tabular}}
		
	\end{table*}
	
	\subsection{Implementation Details}
	The proposed Contextual Mutual Boosting Network (CMBN) is based on the ResNet50 architecture. We fine-tune them with the AMSGrade on the Re-ID datasets. The learning rate is initialized as 0.0003 and decayed by 0.1 every 10 epochs. Person matching is based on the $l_2$ distance of the final representations. The batch size is set to 32. We freeze the ImageNet pre-trained network and train only the FC layers (for feature compression and ID classifier) during the first 5 epochs. Images are resized to 384$\times$128. We only adopt the random flip for data augmentation. We use one NVIDIA RTX-2080Ti GPU for the training and testing, and implement all code on the Pytorch platform.

	\subsection{Visualization of the Activation Maps}
	In Figure 4, we visualize the activation maps from the Backbone Branch. For each example, the activation maps of the IDE and our CMBN are illustrated respectively.
	The IDE has two obvious shortcomings: (1) the salient features are not consistent, many potentially informative cues are ignored, (2) distracting factors like cluttering background pollute the representations enormously. Our proposed CMBN, especially the Reversal Branch and the Contextual-Detachment Strategy, effectively ameliorates the aforementioned two problems.

	\subsection{Comparison with the state-of-the-art}
	To demonstrate the effectiveness of the proposed Contextual Mutual Boosting Network (CMBN), we compare our architecture with the state-of-the-art methods on the Market-1501, DukeMTMC-ReID, and CUHK03-NP datasets and the results are present in Table 1, 2, 3, respectively. Noted that, we compare methods based on the original classification network like ResNet rather than other superior modified networks like PCBNet\cite{sun2018beyond}.

	\textbf{Market-1501.} Compared with the recently proposed attention based method, MGCAM\cite{song2018mask}, HA-CNN\cite{li2018harmonious}, CASN(IDE)\cite{zheng2019re}, our CMBN achieves improvements of at least 0.8\% and 1.8\% on Rank-1 and mAP, respectively. Compared with the segmentation based method \cite{song2018mask} and parsing based method \cite{kalayeh2018human}, we also achieve comparable performance. As for the pose guided methods, PDC \cite{su2017pose}, PN-GAN\cite{qian2018pose}, PGFA\cite{miao2019pose}, our approach performs much better with less computation.

	\textbf{DukeMTMC-ReID.} Again, compared with the attention methods, the Rank-1 and mAP are at least 0.3\% and 2.5\% higher. Compared with the parsing based method SPReID \cite{kalayeh2018human}, it achieves better performance, since the great power of segment person from the background. Compared with the pose guided methods, the Rank-1 and mAP are at least 2.2\% and 4.0\% percent higher.

	\textbf{CUHK03-NP.} Compared with the former two datasets, CUHK03-NP has less training data. On this dataset, we achieve much better results than others on both detected and labeled images. The Rank-1 and mAP are at least 5.7\% and 8.2\% higher on CUHK03-NP Detected and 8.6\% and 10.8\% higher on CUHK03-NP Labeled. It demonstrates that the CMBN can perform well with a small amount of training data.

	Compared with the mentioned networks equipped with various assistances, our network, whose MG and RB components are only auxiliaries during training, requires less computation during the inference stage. Besides, RB and MG are flexible in deploying on other classification architectures, making our network more practical.

	\subsection{Component Analysis}

In this section, we further verify the effectiveness of the proposed `Reversal Branch (RB)', `Mask Generator (MG)', and the `Contextual-Detachment Strategy (CDS)' respectively. In Table 4, we report the evaluation results of the proposed model on CUHK03-NP, Market-1501, and DukeMTMC-ReID, and work up to the final CMBN step by step. Here, We adopt the modified ResNet50 with two additional FC layers (IDE) as the `BaseLine'. From Table 4, we can see clear improvements in performance over the `BaseLine' with the proposed models. 

For instance on the CUHK03-NP (Labeled) dataset, after adding the `RB', the Rank-1 and mAP improves by 7.3\% and 8.8\%, respectively. This provides that localizing pedestrians through mutual contextual boosting does improve the discrimination of representations. It's an important and integral part of our proposed CMBN. Furthermore, we also test adding the `MG' and there are also 11.7\% and 13.6\% improvements of Rank-1 and mAP, respectively. It demonstrates that the channel mask is effective in recalibrating the obtained features. However, after combing the `RB' and `MG' together, the accuracies do not increase but drop compared with the `BaseLine+MG'. 
It means that directly combing the `RB' and `MG' together will pollute the representations since the noise is introduced into the BB from the shared frontend.
Therefore, it is necessary to enforce the foreground learning constraints to the BB especially. 
After adding the `CDS', the Rank-1 and mAP both reach the highest points, proving the importance of the alternately training strategy.

	\section{Conclusion}
    In this paper, we propose the Contextual Mutual Boosting Network, which aims to generate discriminative features of pedestrians for identification. 
    We construct the Reversal Branch, which aims to learn the background information, into the network.
   Through the mutual contextual boosting, the network localizes the spatial regions of pedestrians more accurately.
   Besides, we propose a unique Mask Generator for feature recalibration on the channel dimension by exploiting the statistical inference. At last, we propose the Contextual-Detachment Strategy for optimizing the network. It optimizes the network by learning the multiply contextual information first and then refining the foreground features only. In the inference phase, the Reversal Branch and Mask Generator will be removed, only the shared convolutional frontend, the Backbone Branch, and the generated channel mask are needed, which makes our network more computationally efficient and practical.
   Extensive experiments on the benchmark Re-ID datasets demonstrate the competitive performance compared with other state-of-the-art methods.

	%\newpage
	\bibliographystyle{named}
	\bibliography{ijcai20}
	
\end{document}